\documentclass[dvipsnames]{article}
\usepackage[final]{lychee}
\usepackage[utf8]{inputenc} 
\usepackage[T1]{fontenc}    
\usepackage{url}            
\usepackage{booktabs}       
\usepackage{amsfonts}       
\usepackage{nicefrac}       
\usepackage{booktabs} 
\usepackage{multirow}
\usepackage{wrapfig}
\usepackage{amssymb}
\usepackage{epigraph}
\usepackage{makecell}
\usepackage{listings}
\usepackage{subcaption}
\usepackage{adjustbox}
\usepackage[labelfont=bf]{caption} 
\usepackage{xspace}
\newcommand{\speedup}[1]{\textcolor{blue}{\scriptsize$_{#1\times}$}}
\usepackage{xcolor}
\usepackage[most]{tcolorbox} 
\usepackage{enumitem} 

\definecolor{seedblue}{RGB}{64, 112, 176}
\tcbset{
    casebox/.style={
        colback=seedblue!10!white,
        colframe=seedblue,
        width=\linewidth,
        arc=3mm,
        boxrule=1pt,
        fontupper=\footnotesize,
        fonttitle=\bfseries\small,
        left=2mm, right=2mm, top=2mm, bottom=2mm,
        title style={fill=seedblue},
        breakable 
    }
}
\usepackage{algorithm}
\usepackage{algpseudocode}
\usepackage{color, soul}
\usepackage{microtype}      
\usepackage{xcolor}         
\usepackage{geometry}
\usepackage{times}
\usepackage{ragged2e}
\usepackage{amsmath}
\usepackage{bm}
\usepackage{latexsym}
\usepackage{graphicx}
\usepackage{float}
\usepackage{longtable}
\usepackage{booktabs} 
\usepackage{multirow}
\usepackage{amssymb}
\usepackage{longtable}
\usepackage{tabu}
\usepackage{bbding}
\usepackage{awesomebox}
\usepackage{bbding}
\usepackage{etoolbox}
\usepackage{fancyhdr}
\usepackage{lipsum}
\usepackage{CJKutf8}
\usepackage{pdfpages}
\usepackage{multicol}
\usepackage{pgffor} 
\usepackage{fontawesome5}
\usepackage{nicematrix} 
\usepackage[edges]{forest}
\usepackage{siunitx}
\usepackage{tikz-dependency}
\usepackage{tikz}
\usepackage{bbm}
\usepackage{amssymb}
\usepackage[english]{babel}  
\usepackage{hyphenat}
\usepackage{siunitx}

\newcommand{\Rmnum}[1]{\expandafter\@slowromancap\romannumeral #1@}

\usepackage{caption} 
\usepackage{chngcntr}

\usepackage[colorlinks, linkcolor=RoyalBlue, anchorcolor=BrickRed, citecolor=RoyalBlue, urlcolor=RoyalBlue]{hyperref}

\usepackage{cleveref}
\crefname{section}{§}{§§}
\Crefname{section}{§}{§§}

\newcommand\refsec[1]{Section~\hyperref[sec:#1]{\ref{sec:#1}}}
\newcommand\refsecs[2]{\hyperref[sec:#1]{§\ref{sec:#1}:~\textsc{#1}}, \hyperref[sec:#2]{§\ref{sec:#2}:~\textsc{#2}}}

\usepackage[tikz]{bclogo}
\definecolor{msftBlue}{RGB}{0,164,239}
\definecolor{msftGreen}{RGB}{127,186,0}
\definecolor{msftYello}{RGB}{255,185,0}
\definecolor{mypurple}{RGB}{138,43,226} 
\definecolor{msftBlack}{RGB}{0,0,0}

\newcommand{\ours}{LycheeMemory\xspace}

\newtcolorbox{myboxnote}[1][]{
  breakable,
  title=#1,
  colback=cyan!0,
  colbacktitle=cyan!0,
  coltitle=black,
  fonttitle=\bfseries,
  bottomrule=0pt,
  toprule=0pt,
  leftrule=1.5pt,
  rightrule=1.5pt,
  titlerule=0pt,
  arc=0pt,
  outer arc=0pt,
  colframe=lightgray,
}
\usepackage{mdframed}

\definecolor{academicblue}{RGB}{54, 95, 145}

\tcbset{
  iclrtakeawaybox/.style={
    width=\linewidth,
    colback=gray!5,                 
    colframe=gray!60,              
    colbacktitle=gray!30,            
    coltitle=black,                
    fonttitle=\bfseries,     
    boxrule=0.5pt,                 
    arc=2pt,                       
    sharp corners=south,           
    attach boxed title to top left={yshift=-0.1in,xshift=0.15in},
    boxed title style={boxrule=0pt, colframe=white},
    enhanced,
    drop shadow southeast           
  }
}
\newtcolorbox{TakeawayBox}[2][]{iclrtakeawaybox,title=#2,#1}

\usepackage{fancyhdr} 
\usepackage{blindtext} 
\usepackage{ulem}

\usepackage{xparse}
\usepackage{colortbl}

\usepackage{svg}

\title{
\vspace{-2em}
\fontsize{16}{19}\selectfont Dynamic Long Context Reasoning over Compressed Memory via End-to-End Reinforcement Learning}
\author{
 Zhuoen Chen, Dongfang Li, Meishan Zhang, Baotian Hu, Min Zhang \\
Research Institute of Computing and Intelligence\\
{Harbin Institute of Technology, Shenzhen}\\
}

\begin{document}

\maketitle
\vspace{-1em}

\begin{abstract}
Large Language Models (LLMs) face severe challenges in long-context processing, including quadratic computational costs, information forgetting, and the context fragmentation inherent in Retrieval-Augmented Generation (RAG). 
We introduce {\ours}, a cognitively inspired framework that enables efficient long-context inference via chunk-wise compression and selective memory recall, rather than processing all raw tokens.
{\ours} segments the input into chunks and encodes each into compressed KV-cache-style representations using a \texttt{Compressor}. A \texttt{Gate} then dynamically selects relevant memory blocks, which a \texttt{Reasoner} iteratively processes with an evolving working memory to solve downstream tasks.
The \texttt{Compressor} and \texttt{Reasoner} are jointly optimized via end-to-end reinforcement learning, while the \texttt{Gate} is trained separately as a classifier.
Experimental results demonstrate that {\ours} achieves competitive accuracy (up to 82\% in ablation variants) on multi-hop reasoning benchmarks (e.g., RULER-HQA), successfully extrapolates context length from 7K to 1.75M, and provides a favorable accuracy--efficiency trade-off against strong long-context baselines. Notably, compared to MemAgent, {\ours} achieves an average 2$\times$ reduction in peak GPU memory usage and a 6$\times$ speedup during inference.
\end{abstract}

\vspace{1em} 
\begin{center}
    \centering
    \includegraphics[width=0.95\textwidth]{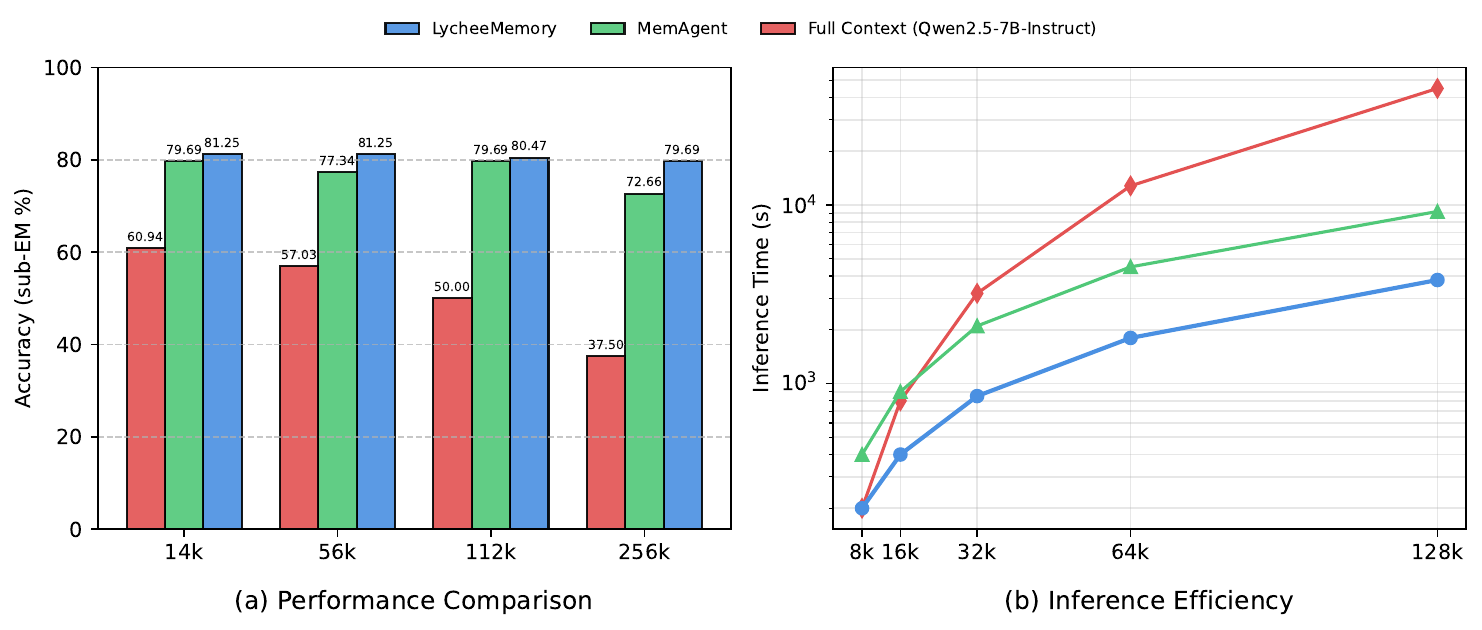}
    \vspace{0.5em}
    \captionof{figure}{\ours achieved the best performance and latency. 
    \textbf{Left:} Relative performance comparison of various methods on the Qwen2.5-7B model across different LongBench datasets. 
    \textbf{Right:} Inference time comparison across different context lengths of 128 samples.}
    \label{fig:teaser}
\end{center}
\vspace{2em} 


\newpage
{
  \hypersetup{linkcolor=RoyalBlue, linktoc=page}
  \tableofcontents
}

\clearpage

\section{Introduction}

Despite the remarkable capabilities demonstrated by Large Language Models (LLMs), efficiently processing long contexts remains a critical challenge~\cite{liu2025comprehensive,comanici2025gemini,wan2025qwenlong}. To address this bottleneck, current methodologies primarily diverge into three paradigms, each facing inherent trade-offs between efficiency and capability. Sparse and linear attention mechanisms~\citep{beltagy2020longformer, xiao2023efficient, katharopoulos2020transformers} reduce computational complexity but often suffer from performance degradation on extremely long sequences. Retrieval-Augmented Generation (RAG)~\citep{lewis2020retrieval, karpukhin2020dense, izacard2021leveraging} mitigates length constraints while facing severe context fragmentation. By treating text chunks as independent entities, it disrupts the logical dependencies essential for multi-hop reasoning and struggles to capture implicit semantic connections. Conversely, recurrent architectures like RecurrentGPT~\citep{zhou2023recurrentgpt} and MemAgent~\cite{yu2025memagent} rely on sequential state updates, resulting in slow serial inference speeds that significantly hinder scalability.


To overcome these limitations, we draw inspiration from the mechanisms of human memory and propose~\ours. By mimicking the division of labor between compressed memory bank (i.e., \textit{long-term memory}) and dynamic working memory~\cite{atkinson1968human}, our framework splits the input text into chunks and compresses them into efficient, high-fidelity compressed KV-cache representations. This builds a compressed memory bank that preserves semantic information while reducing computational costs.
During inference, we use a dynamic recall and reasoning workflow driven by a \texttt{Gate} and a \texttt{Reasoner}. It starts with an empty \textit{working memory}, explicitly instantiated as a fixed-length, token-level context window. This design preserves a discrete action space, thereby enabling the memory update process to be optimized via Reinforcement Learning (RL). Subsequently, \ours sequentially traverses the compressed memory bank: for each chunk, the \texttt{Gate} evaluates whether the chunk contributes to the current reasoning state, given the current working memory and the user query. If deemed relevant, the \texttt{Reasoner} utilizes the chunk to update current working memory; otherwise, the chunk is skipped. Through this selective update and iterative refinement, the \texttt{Reasoner} facilitates multi-step reasoning across multiple memory chunks, avoiding the blind processing of the entire input sequence characteristic of traditional recurrent architectures.

A core challenge is ensuring that the compressed memory can be effectively used by the \texttt{Reasoner} for precise inference. We adopt a joint policy optimization strategy: we train the \texttt{Compressor} and \texttt{Reasoner} end-to-end with RL, and train the \texttt{Gate} separately as a classifier.
We evaluate \ours on RULER-HQA~\cite{yang2018hotpotqa, hsiehruler}, 2WikiMultihopQA~\cite{ho2020constructing}, and StreamingQA~\cite{liska2022streamingqa}. 
Experimental results show that {\ours} maintains competitive accuracy on multi-hop reasoning, extrapolates context length from 7K to 1.75M, and improves the accuracy--efficiency trade-off. Compared to MemAgent~\cite{yu2025memagent}, {\ours} reduces peak GPU memory usage by 2$\times$ and speeds up inference by 6$\times$.

The main contributions of this work are summarized as follows:
\begin{itemize}
    \item We propose \ours, a framework comprising a \texttt{Compressor}, a \texttt{Gate}, and a \texttt{Reasoner}, which transforms long-context processing from direct modeling of raw tokens into efficient iterative reasoning over a compressed memory bank.
    \item We introduce a joint policy optimization strategy that trains the \texttt{Compressor} and \texttt{Reasoner} end-to-end via RL, enabling the compressed memory to be directly optimized for downstream tasks.
    \item Experimental results show that \ours scales the context size to 1.75M tokens and improves inference efficiency while maintaining competitive accuracy.
\end{itemize}
\section{Related Work}
\label{sec:related_work}


\paragraph{Explicit Memory Methods.}
Explicit memory methods externalize context as human-readable text or symbols. 
Standard RAG retrieves static chunks via semantic similarity but often suffers from context fragmentation and limited precision in multi-hop reasoning~\cite{gutierrez2025from,weller2025theoreticallimitationsembeddingbasedretrieval,merola2025reconstructingcontextevaluatingadvanced}. 
Agentic memory systems mitigate this by actively managing external memory, such as MemGPT’s OS-inspired hierarchy~\cite{DBLP:journals/corr/abs-2310-08560} and Mem0’s lifecycle-based memory updates~\cite{chhikara2025mem0buildingproductionreadyai}. 
More recent RL-based approaches (e.g., MemAgent~\cite{yu2025memagent}, Mem1~\cite{zhou2025mem1learningsynergizememory}) learn to manage a bounded memory by selectively overwriting or integrating observations during streaming.
Despite their interpretability, these methods operate on raw tokens and incur substantial computational overhead.
In contrast, our approach leverages compressed memory with selective retrieval, achieving lower peak memory usage and inference latency.

\paragraph{Implicit Memory Methods.}
Implicit memory methods optimize internal representations via activation compression or parametric updates. 
To alleviate the quadratic cost of self-attention, cache compression approaches exploit attention sparsity, retaining only salient tokens (e.g., H2O~\cite{zhang2023ho}, SnapKV~\cite{li2024snapkv}). 
Beyond static pruning, dynamic methods retrieve relevant cache blocks on demand~\cite{xiao2024infllm,gao2025quest}. 
Parametric alternatives, such as DyPRAG~\cite{tan2025dynamicparametricretrievalaugmented}, encode documents into latent LoRA adapters~\cite{hu2022lora} and route queries to specialized weights. 
While effective in reducing memory footprint, aggressive compression often degrades long-tail reasoning~\cite{zhang2025long}, and purely latent approaches~\cite{DBLP:journals/corr/abs-2412-06769,eyuboglu2025cartridges} lack structured retrieval needed for large-scale multi-document streams. 
In contrast, our method couples selective retrieval with iterative working memory updates via a \texttt{Gate} and \texttt{Reasoner}, enabling robust multi-hop reasoning over million-token contexts.

Overall, \ours bridges explicit and implicit memory: it stores documents as compressed KV-cache representations, while performing state-dependent retrieval and reasoning through a plaintext working memory. This retains the scalability benefits of compression and yields an interpretable trace over selected evidence chunks.

\begin{figure*}[t]
  \centering
  \includegraphics[width=\textwidth]{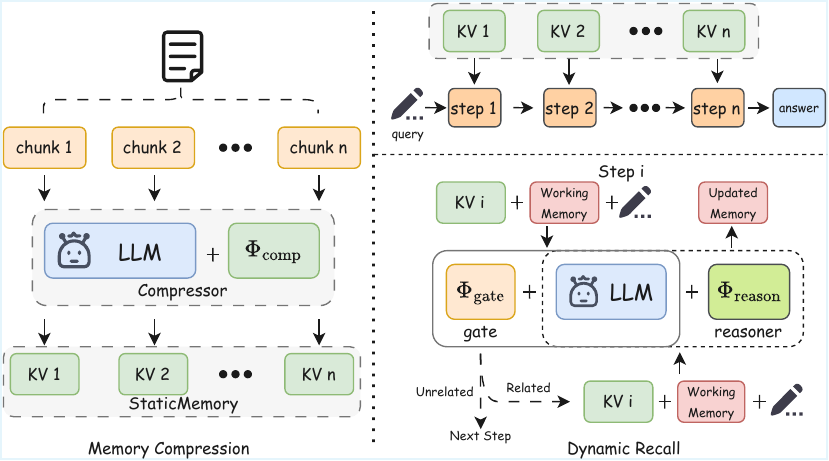}
  \caption{Overview of the {\ours} framework. The left panel illustrates compressed memory construction, where a long document is segmented and compressed into compact KV-cache representations by the compressor. The right panel depicts the dynamic recall and reasoning workflow, in which the gate selectively activates relevant memory blocks and the reasoner iteratively updates the working memory to produce the final answer.}
  \label{fig:architecture}
  \vspace{-4mm}
\end{figure*}
\section{Methodology}
\label{sec:methodology}

\subsection{Overview}
\label{sec:task_formulation}
We address long-context modeling where a model takes ultra-long documents $D$ (length $N$) and a user query $Q$ to generate an answer $A$. Due to the prohibitive length of $D$, processing the entire sequence directly is computationally infeasible. 
To address this, we propose \textbf{{\ours}}, a dual-system framework for long-context processing. As illustrated in Figure \ref{fig:architecture}, the architecture comprises three core roles:
\begin{itemize}
    \item \texttt{Compressor} $\Phi_{\text{comp}}$: Composed of the base model $\Phi$ augmented with a compression LoRA module $\Psi_{\text{comp}}$, responsible for encoding raw text into KV-cache-style memory.
    \item \texttt{Gate} $\Phi_{\text{gate}}$: Composed of the base model $\Phi$ augmented with a gating LoRA module $\Psi_{\text{gate}}$, acting as a relevance filter.
    \item \texttt{Reasoner} $\Phi_{\text{reason}}$: Composed of the base model $\Phi$ augmented with a reasoning LoRA module $\Psi_{\text{reason}}$, responsible for complex reasoning based on recalled memories.
\end{itemize}
Let $D$ be segmented into $K$ sequential chunks (size $sz$, i.e., $K=N/sz$) as $D = \{C_1, C_2, \dots, C_K\}$. In our experiments, we set $sz=4096$.
The processing workflow of {\ours} involves two main phases: 

\paragraph{Memory Compression:} In this phase (detailed in \S\ref{sec:static_memory}), each text chunk $C_k$ is processed by the \texttt{Compressor} $\Phi_{\text{comp}}$ and encoded into a compact latent representation $\theta_k$. This representation is subsequently stored in the compressed memory bank $\Theta$, i.e., $\Theta = \{\theta_1, \dots, \theta_K\}$. 

\paragraph{Dynamic Recall and Reasoning:} Distinct from the latent representations used for storage, the model maintains a working memory $\mathbf{m}$ during the dynamic recall and reasoning phase. $\mathbf{m}$ exists as plaintext tokens within the model's context and is iteratively updated as the model scans the compressed memory bank to maximize reasoning capability. When receiving a user query $Q$ (detailed in \S\ref{sec:dynamic_recall}), the model scans memory blocks with index $i=1,\dots,K$, activating the \texttt{Gate} $\Phi_{\text{gate}}$ and \texttt{Reasoner} $\Phi_{\text{reason}}$. The process starts with an initial empty working memory $\mathbf{m}_0$. At scan step $i$, the \texttt{Gate} evaluates the compressed memory block $\theta_i$ in conjunction with the current working memory $\mathbf{m}_t$ and query $Q$. If deemed relevant, the \texttt{Reasoner} is invoked to update the working memory state: $\mathbf{m}_{t+1} = \Phi_{\text{reason}}(\mathbf{m}_t, \theta_i, Q)$, and we increment the update index $t \leftarrow t+1$; otherwise, we skip this block and keep $t$ unchanged. Finally, the model synthesizes the answer $A$ based on $\mathbf{m}_T$ and $Q$, where $T \le K$.

\subsection{Compressed Memory Construction}
\label{sec:static_memory}
The construction of the compressed memory bank $\Theta$ is central to the {\ours} framework. We present a KV-cache compression style method that achieves an optimal balance between information density and computational efficiency.

\subsubsection{KV-cache Style Compression via Memory Tokens}
\label{sec:kv_compression} 
Similar to previous works~\cite{chevalier2023adapting,deng2025unigist}, we define the compression as a mapping from text to a latent representation, $C_i \rightarrow \theta_i$. We utilize base model $\Phi$ augmented by a LoRA module $\Psi_{\text{comp}}$ as the \texttt{Compressor}, eliminating the need for an external encoder.
For any text chunk $C_i = [x_1^i, \dots, x_w^i]$ of length $w$, we first determine a compression ratio $\alpha_i$. We then define a set of $z_i = w / \alpha_i$ trainable memory tokens $V_i = \{\langle v \rangle_1^i, \dots, \langle v \rangle_{z_i}^i\}$.
Next, we interleave $V_i$ with $C_i$ by inserting a memory token after every $\alpha_i$ original tokens, forming interleaved sequence $C'_i$:
\begin{equation*}
\begin{aligned}
C'_i
&= \operatorname{Interleave}(C_i, V_i) = [x_1^i, \dots, x_{\alpha_i}^i, \langle v\rangle_1^i, \dots, x_w^i, \langle v\rangle_{z_i}^i]
\end{aligned}
\end{equation*}
This sequence $C'_i$ is passed through $\Phi_{\text{comp}}$ for a single forward pass. During this process, the model is trained to embed the semantic information of the preceding $\alpha_i$ tokens into the hidden state of the subsequent memory token $\langle v \rangle_j^i$.
Finally, the set of hidden states corresponding to all memory tokens constitutes the compact KV-cache style representation $\theta_i$ stored in the compressed memory bank $\Theta$:
\begin{align*}
& \theta_i \ = \{ h(\langle v \rangle_1^i), \dots, h(\langle v \rangle_{z_i}^i) \}, \\
& \text{where } \ h(\cdot) = \operatorname{HiddenState}\!\left( \Phi_{\text{comp}}(C'_i) \right) .
\end{align*}

\subsubsection{Pre-optimization of the Compressor}
\label{sec:compression_training}
Before end-to-end RL, to ensure that $\theta_i$ retains the core semantic information of $C_i$ despite high compression, we jointly optimize the LoRA module $\Psi_{\text{comp}}$ while keeping the base model $\Phi$ frozen using data augmentation and diverse tasks.
Note that in the encoding phase ($C'_i \rightarrow \theta_i$), the base model $\Phi$ combined with $\Psi_{\text{comp}}$ generates the compressed representation $\theta_i$. Conversely, in all subsequent decoding tasks based on $\theta_i$, we utilize only the frozen base model $\Phi$ without $\Psi_{\text{comp}}$ for generation. This design ensures that the gradient flows only through $\Psi_{\text{comp}}$, effectively decoupling the compression capability from the general generation ability of the base model.
Given a compressed representation $\theta_i$, the model $\Phi$ is trained to perform three distinct tasks. Let $P_{\Phi}(Y | \text{context})$ be the probability generating $Y$ given the context:

\paragraph{Text Reconstruction.} The model must regenerate the original text $C_i$ using only $\theta_i$ as context.
\[\mathcal{L}_{\text{recon}} = - \log P_{\Phi}(C_i | \theta_i)\]

\paragraph{QA Generation.} We pre-generate synthetic question-answer pairs $(Q_j, A_j)$ for $C_i$. The model generates $A_j$ given $\theta_i$ and $Q_j$. The loss $\mathcal{L}_{\text{qa}}$ is computed only over the answer $A_j$.
\[\mathcal{L}_{\text{qa}} = - \mathbb{E}_{(Q_j, A_j) \sim C_i} [\log P_{\Phi}(A_j | \theta_i, Q_j)]\]

\paragraph{Creative Generation.} The model performs high-level semantic tasks based on $\theta_i$, such as generating a summary $S_i$. We use the model output based on the original text, $\Phi(C_i)$, as the ground-truth label $Y_{\text{creative}}$.
\begin{align*}
    & \mathcal{L}_{\text{creat}} = - \log P_{\Phi}(Y_{\text{creat}} | \theta_i) \quad \\
    & \text{where } Y_{\text{creat}} = \Phi(C_i, P_{\text{task}})
\end{align*}
The total loss $\mathcal{L}_{\text{comp}}$ is a weighted sum of the above losses, minimized by updating $\Psi_{\text{comp}}$:
\[
\min_{\Psi_{\text{comp}}} \mathcal{L}_{\text{comp}} = \mathbb{E}_{C_i \sim \mathcal{D}} [ w_1 \mathcal{L}_{\text{recon}} + w_2 \mathcal{L}_{\text{qa}} + w_3 \mathcal{L}_{\text{creat}} ]
\]
We train separate projection matrices for the memory tokens $v_{\text{mem}}$, functionally isolating them from regular token representations to learn a dedicated compression subspace.

\subsection{Dynamic Recall and Reasoning Workflow}
\label{sec:dynamic_recall}

After constructing the compressed memory bank $\Theta$, the core of {\ours} lies in efficiently retrieving and reasoning over these compressed representations. In contrast to methods like MemAgent~\citep{yu2025memagent}, which employ linear scanning with forced updates for every chunk, we introduce a relevance threshold $\tau$. As the system traverses the compressed memory bank, the \texttt{Gate} scores each compressed memory block, and only blocks exceeding this threshold trigger the \texttt{Reasoner} to update the working memory.

\subsubsection{LoRA Gate}
\label{sec:gating}
To avoid the overhead of unnecessary memory updates, we require a filter to discard static blocks irrelevant to the user query $Q$. An intuitive solution would be an embedding model calculating cosine similarity between chunks and $Q$. While such lightweight retrieval can be reasonably strong on recall, it only captures static semantic similarity and lacks state-dependent retrieval conditioned on the evolving working memory $\mathbf{m}$ (See \S\ref{sec:experiments}). This limitation becomes salient in multi-hop settings, where later-hop evidence may only become relevant after intermediate entities are added into $\mathbf{m}$. A further critical limitation is that external retrievers cannot leverage the working memory $\mathbf{m}$, which often contains key secondary clues (e.g., intermediate entities) derived from the query and previously processed memory chunks.
Motivated by this, we implement the \texttt{Gate} $\Phi_{\text{gate}}$ by adding a LoRA adapter $\Psi_{\text{gate}}$ to the base model $\Phi$.

\paragraph{Architecture and Inference.}
Given a user query $Q$, the current working memory $\mathbf{m}_t$, and a candidate memory block $\theta_i \in \Theta$ (represented by its memory tokens), we concatenate them and extract the hidden state of the final token, $\mathbf{h}_{\text{last}}$. This state is projected by a trainable linear head $\mathbf{W}_{\text{gate}}$ followed by a sigmoid activation to produce a relevance probability:
\[
P = \sigma(\mathbf{W}_{\text{gate}} \cdot \mathbf{h}_{\text{last}}(\Phi(Q, \mathbf{m}_t, \theta_i; \Psi_{\text{gate}})))
\]
The memory block is used to update the working memory only if $P > \tau$.

\paragraph{Training Objective.}
Due to the gradient discontinuity caused by discrete recall decisions, we treat gating training as a separate binary classification task beyond RL in \S\ref{sec:rl_optimization}. We align text chunks with downstream tasks (e.g., QA pairs) to construct training data. A memory block $\theta_i$ is labeled positive ($y_i^*=1$) if it contains evidence required to answer $Q$, and negative ($y_i^*=0$) otherwise. We optimize the gate parameters (LoRA $\Psi_{\text{gate}}$ and Head $\mathbf{W}_{\text{gate}}$) using Binary Cross-Entropy (BCE) loss:
\begin{align*}
\mathcal{L}_{\text{gate}}
&= - \frac{1}{N} \sum_{i=1}^{N} \Big[y_i^* \log P_i + (1 - y_i^*) \log (1 - P_i)\Big]
\end{align*}
where $P_i$ is the predicted probability. This lightweight design ensures the model identifies memory chunks relevant to the query and current reasoning state in the latent space.

\subsection{End-to-End RL Optimization}
\label{sec:rl_optimization}

To empower {\ours} with the capability of complex reasoning over compressed memories, we propose an enhanced reinforcement learning framework. Unlike prior approaches that optimize components in isolation, we formulate the entire lifecycle from memory compression to reasoning as a unified joint policy optimization problem. This allows the gradient from the final reasoning outcome to backpropagate through the recall workflow and update the \texttt{Compressor}, ensuring the $\Theta$ (i.e., \textit{long-term memory}) is optimized specifically for downstream inference.

\subsubsection{Joint Policy Formulation}
We define the joint policy $\pi_\vartheta$ parameterized by $\vartheta$, which encompasses both the \texttt{Compressor} parameters ($\Psi_{\text{comp}}$) and the \texttt{Reasoner} parameters ($\Psi_{\text{reason}}$).
For a given input document $D$ and query $Q$, the generation of an answer $A$ involves a hierarchical trajectory:
\begin{align*}
\pi_\vartheta(A, \mathcal{M}, \Theta \mid D, Q)
&=
\underbrace{\prod_{k=1}^K \pi_{\text{comp}}(\theta_k \mid C_k)}_{\text{Memory Construction}}
\cdot\!\underbrace{\prod_{t=1}^T \pi_{\text{reason}}(\mathbf{m}_t \mid \mathbf{m}_{t-1}, \Theta, Q)}_{\text{Dynamic Recall and Reasoning}}
\end{align*}
where $\Theta = \{\theta_k\}$ represents the compressed memory bank, and $\mathcal{M} = \{\mathbf{m}_t\}_{t=0}^{T}$ represents the sequence of working memory updates. Our goal is to maximize the expected reward of the final answer $A$ by optimizing $\vartheta$.

\paragraph{The Unified Objective Function.}
We formulate the unified objective to jointly optimize compression and reasoning:
\begin{align*}
& \mathcal{J}(\vartheta) = \mathbb{E}_{Q \sim \mathcal{D}, \{O_i\}_{i=1}^G \sim \pi_{\vartheta_{\text{old}}}} 
\left[ \frac{1}{G} \sum_{i=1}^G \frac{1}{n_i} \sum_{j=1}^{n_i} \left( \mathcal{L}^{\text{CLIP}}_{i,j}(\vartheta) - \beta D_{\text{KL}}(\pi_\vartheta \| \pi_{\text{ref}}) \right) \right] \\
&\text{where} \quad \mathcal{L}^{\text{CLIP}}_{i,j}(\vartheta) = 
\min \left( \rho_{i,j}(\vartheta) \hat{A}_{i,j}, \, \text{clip}(\rho_{i,j}(\vartheta), 1-\epsilon, 1+\epsilon) \hat{A}_{i,j} \right)
\end{align*}
Here, $\rho_{i,j}(\vartheta)$ represents the sequence-level importance sampling weight defined by GSPO~\cite{zheng2025group}. $G$ denotes the group size (number of sampled trajectories per prompt), and $n_i$ denotes the number of tokens in the $i$-th trajectory. By maximizing $\mathcal{J}(\vartheta)$, the model learns to compress context into $\Theta$ such that the reasoning policy maximizes the likelihood of high-advantage trajectories.

\subsection{Complexity and Efficiency Analysis}
\label{sec:complexity}
In this section, we analyze the computational efficiency of {\ours} compared to existing long-context methods. 

\paragraph{Memory Construction.} This phase incurs $O(N)$ complexity, but it is a one-time, fully parallelizable pre-processing cost.

\paragraph{Gate.}
While approaches like MemAgent~\citep{yu2025memagent} achieve $O(N/sz)$ linear complexity via streaming, they require performing full token generation (i.e., memory updates) for every text chunk. 
In contrast, although the \texttt{Gate} in {\ours} must also traverse all $K$ compressed memory blocks to determine relevance, maintaining an $O(N/sz)$ complexity, the computational cost per block is drastically reduced. The \texttt{Gate} requires only a single forward pass for scalar classification, rather than the computationally expensive autoregressive generation used in standard streaming methods.

\paragraph{Dynamic Recall and Reasoning.}
The heavy computational load of the \texttt{Reasoner} is decoupled from the document length $N$ and depends only on the number of retrieved blocks $T$:
\begin{align*}
\mathcal{O}_{\text{inference}} \approx O(N /sz \times C_{\text{gate}} + T \times C_{\text{reason}})
\end{align*}
where $C_{\text{gate}} \ll C_{\text{reason}}$. Since the \texttt{Gate} efficiently filters out irrelevant information ($T \ll N / sz$), {\ours} achieves a significantly lower constant factor in its linear scaling compared to methods that reason over every chunk.

\section{Experiments}
\label{sec:experiments}
In this section, we evaluate {\ours} on long-context QA, analyze inference efficiency and zero-shot generalization, and validate core design choices through ablations.

\subsection{Experimental Setup}
\label{sec:exp_setup}
\paragraph{Model Configuration} We use Qwen2.5-Instruct~\cite{qwen2025qwen25technicalreport} as the base model and train \ours-3B/\ours-7B initialized from Qwen2.5-3B/7B-Instruct.
\paragraph{Dataset Construction} Following MemAgent~\cite{yu2025memagent}, we synthesize long-document training data from RULER-HQA~\cite{yang2018hotpotqa,hsiehruler} by mixing query-relevant articles with distractors (avg. 20K tokens).
We evaluated contexts from 7K to 1.75M tokens for length extrapolation and reported zero-shot results on 2WikiMultihopQA~\cite{ho2020constructing}, StreamingQA~\cite{liska2022streamingqa}.
\paragraph{Baselines} We compare with Search-R1~\cite{jin2025search}, MemAgent~\cite{yu2025memagent}, DeepSeek-R1-Distill-Qwen~\cite{guo2025deepseek}, Qwen-2.5-Instruct-1M~\cite{yang2025qwen2}, and QwenLong-L1~\cite{wan2025qwenlong}, using official configurations.
Additional details are in Appendix~\ref{app:training_details} and Appendix~\ref{app:eval_baselines}.

\subsection{Main Results}
\label{sec:main_results}
\begin{table*}[t]
  \centering
  \small
  \setlength{\tabcolsep}{3pt}  
  \resizebox{\linewidth}{!}{
    \begin{tabular}{llllllllll}
      \toprule
      \textbf{Model} & \textbf{7K} & \textbf{14K} & \textbf{28K} & \textbf{56K} & \textbf{112K} & \textbf{224K} & \textbf{448K} & \textbf{896K} & \textbf{1.75M}\\
      \midrule
      QwenLong-L1-32B~\cite{wan2025qwenlong} & 72.66 & 75.00 & 72.66 & 60.94 & 31.25 & 17.19 & 13.28 & 11.72 & OOM\\
      Qwen2.5-Instruct-14B-1M~\cite{yang2025qwen2} & 60.16 & 60.94 & 50.00 & 57.03 & 50.00 & 37.50 & 8.59 & 0.00 & OOM\\
      Qwen2.5-Instruct-7B-1M~\cite{yang2025qwen2} & 61.72 & 56.25 & 53.91 & 55.47 & 51.56 & 33.59 & 12.50 & 0.00 & OOM\\
      \midrule
      DS-Distill-Qwen-32B~\cite{guo2025deepseek} & 70.31 & 66.41 & 65.62 & 46.88 & 23.44 & 13.28 & 7.81 & 7.03 & OOM\\
      DS-Distill-Qwen-14B~\cite{guo2025deepseek} & 64.06 & 64.84 & 57.03 & 40.62 & 14.84 & 8.59 & 3.12 & 6.25 &OOM\\
      DS-Distill-Qwen-7B~\cite{guo2025deepseek} & 30.47 & 12.50 & 3.12 & 0.00 & 0.00 & 0.78 & 0.00 & 0.00 & OOM\\
      \midrule
      RAG + Qwen2.5-7B-Instruct & 67.19  & 66.41  & 66.41  & 67.19  & 64.84  & 64.06  & 62.5  & 61.72  & 62.38 \\
      Search-R1~\cite{jin2025search} & 72.66 & 71.88 & 67.71  & 73.96  & 66.67  & 62.5 &  64.58  & 67.71  & 67.19 \\
      \midrule
      RL-MemAgent-7B~\cite{yu2025memagent} & \textbf{82.03} & 79.69 & 78.91 & 77.34 & 79.69 & 72.66 & 74.22 & \textbf{76.56} & 75.78\\
      \midrule
      \textbf{\ours-7B} (ours)
      & 77.34\speedup{1.0}
      & 76.56\speedup{1.2}
      & 75.00\speedup{1.6}
      & 76.56\speedup{2.5}
      & 75.78\speedup{3.5}
      & 73.44\speedup{5.9}
      & 74.22\speedup{9.7}
      & 72.66\speedup{17.7}
      & 71.09\speedup{28.2}
      \\
      \textbf{\ours-7B w/o Gate} (ours)  & 80.47 & \textbf{81.25} & \textbf{82.03} & \textbf{81.25} & \textbf{80.47} & \textbf{79.69} & \textbf{75.00} & 75.78 & \textbf{78.12}\\
      \bottomrule
  \end{tabular}}
  \caption{
    Comparison of main experimental results under different context lengths.
    All values are normalized sub-EM accuracy (\%).
    \textcolor{blue}{Blue} indicates the inference speedup of \ours relative to its without \texttt{Gate} ablation.
  }
  \label{tab:main_results}
\end{table*}

We first evaluate {\ours} on the synthesized HotpotQA dataset as context length grows. Table~\ref{tab:main_results} shows the comparison with baselines.
\paragraph{Performance at Scale} We compare models from 7K to 896K context lengths. For memory-based models (Search-R1, MemAgent, and {\ours}), we further evaluate extrapolation at an ultra-long 1.75M tokens to inspect generalization beyond standard training ranges. As shown in Table~\ref{tab:main_results}, several baselines fail even within their nominal windows. Reasoning models (e.g., DS-Distill-Qwen series) degrade rapidly as context length increases. In contrast, MemAgent and {\ours} show strong length extrapolation, with only mild performance drop as input length increases, validating the effectiveness of the chunked memory mechanism.
\paragraph{Comparison with MemAgent} Compared to MemAgent, our \ours-7B w/o Gate ablation achieves higher accuracy across most evaluated context lengths, while \ours with Gate trades a small accuracy drop for substantially improved inference efficiency (see \S\ref{sec:efficiency}). This indicates that compressed memory with RL-trained reasoning is competitive in accuracy, and the Gate provides an effective accuracy--efficiency trade-off in ultra-long contexts.

\subsection{Inference Efficiency Analysis}
\label{sec:efficiency}
\begin{figure}[t]
  \centering
  \includegraphics[width=1\linewidth]{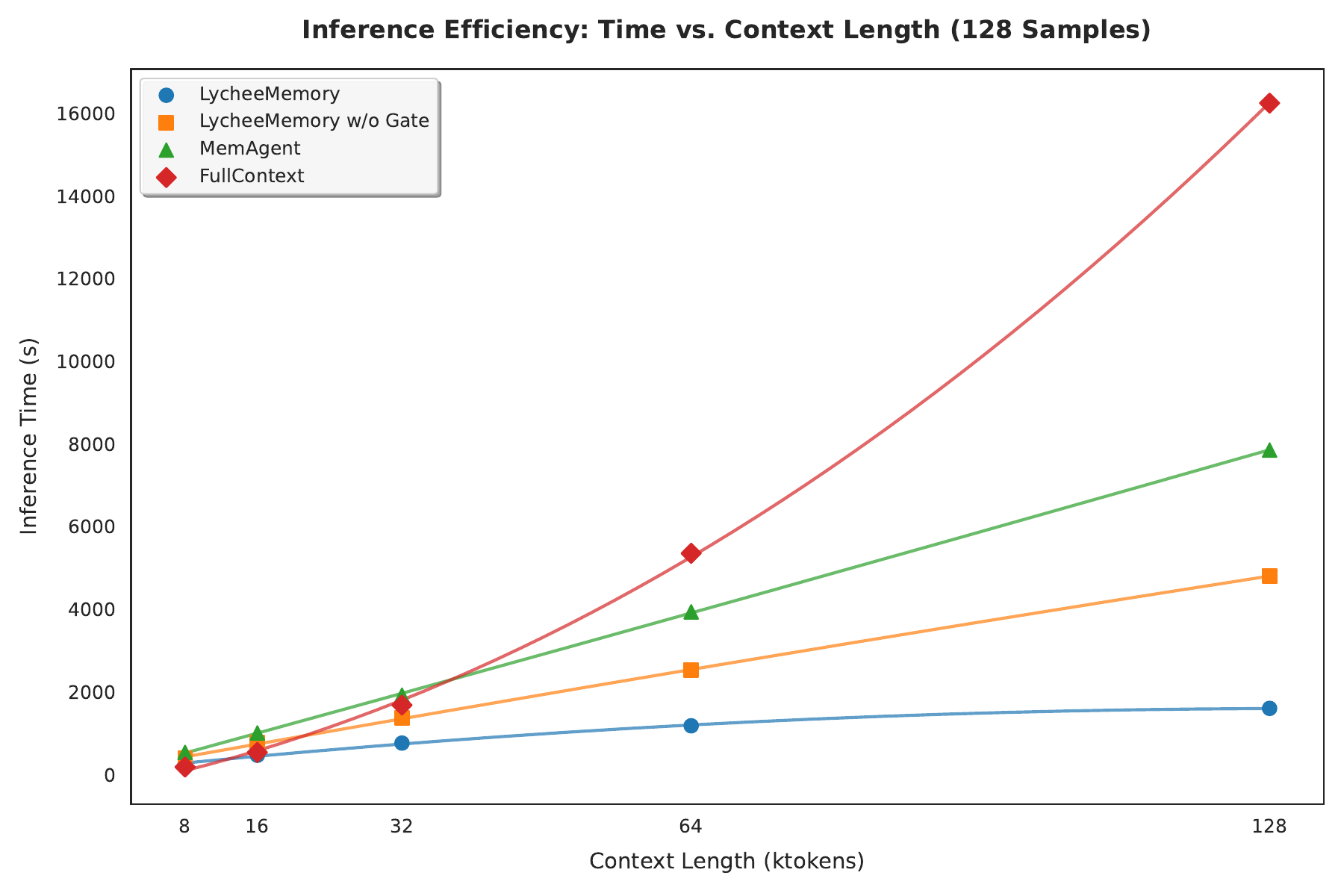}
  \caption{Inference latency as context length increases. {\ours} exhibits a nearly flat latency curve, in contrast to the quadratic and linear increases observed in the full-context and MemAgent baselines respectively.}
  \label{fig:efficiency}
\end{figure}

A key advantage of {\ours} is computational efficiency. We measure end-to-end inference time on 2$\times$ A100 (80GB) for 128 samples from 8K to 128K tokens (generation length 1024, largest non-OOM batch). The reported time includes compression and I/O. Figure~\ref{fig:efficiency} shows three regimes:
\paragraph{Quadratic Explosion} The Qwen2.5-7B baseline exhibits the expected $O(N^2)$ latency growth. At 64K it is markedly slower than memory-based methods and at 128K it further fails due to OOM.
\paragraph{Linear Growth} MemAgent and our ablation {\ours} without \texttt{Gate} (linear scan over all compressed memory blocks) show linear $O(N)$ complexity. However, {\ours} without \texttt{Gate} remains faster than MemAgent because our compressed memory is a highly compressed KV-cache ($\alpha \gg 1$), so the effective sequence length processed by the reasoning workflow is much shorter than the text stream of MemAgent.
\paragraph{Near-Constant Inference} With the \texttt{Gate} module, {\ours} shows striking efficiency. As context grows from 8K to 128K, inference time rises only slightly. Compression and Gate overhead grows linearly (with a tiny coefficient), while the costly reasoning (with memory update) steps run on only a few retrieved blocks. In terms of results, at 128K we achieve a 6$\times$ speedup over MemAgent and a 3.5$\times$ speedup over the w/o Gate baseline; meanwhile, Table~\ref{tab:main_results} shows that the accuracy drop on the closest reported bucket (112K) is only 6\%.
Additional analyses are in Appendix~\ref{app:storage_tradeoff} and Appendix~\ref{app:complexity_analysis}.

\begin{table*}[b]
  \centering
  \small

  \begin{minipage}[t]{0.49\textwidth}
    \vspace{0pt}
    \centering
    \setlength{\tabcolsep}{5pt}
    \resizebox{\linewidth}{!}{
      \begin{tabular}{lccc|cc}
        \toprule
        \textbf{Method} &
        \multicolumn{3}{c|}{\textbf{2WikiMultihopQA}} &
        \multicolumn{2}{c}{\textbf{StreamingQA}} \\
        &
        \textbf{14K} & \textbf{28K} & \textbf{56K} &
        \textbf{F1} & \textbf{sub-EM} \\
        \midrule

        Qwen2.5-Instruct-7B     & 57.0 & 42.2 & 37.5 & 30.5 & 23.4 \\
        RAG          & 68.8 & 64.1 & 59.4 & \textbf{84.3} & 67.2 \\
        MemAgent     & 74.2 & \textbf{73.4} & 71.1 & 77.9 & 60.2 \\
        \textbf{\ours} &
        \textbf{75.0} & 70.3 & \textbf{73.4} &
        80.8 & \textbf{73.4} \\
        \bottomrule
      \end{tabular}}
    \caption{Zero-shot comparison results of 2WikiMultihopQA and StreamingQA.}
    \label{tab:2WikiMultihopQA}
  \end{minipage}\hfill
  \begin{minipage}[t]{0.49\textwidth}
    \vspace{9pt}
    \centering
    \resizebox{\linewidth}{!}{
      \begin{tabular}{lccc}
        \toprule
        \textbf{Method} & \textbf{56K} & \textbf{112K} & \textbf{224K} \\
        \midrule
        Text-embedding-3-large & 94.3 & 82.1 & 80.9 \\
        Gate (Query Only) & 88.2 & 76.4 & 74.8 \\
        \textbf{Gate (Query + Memory)} & \textbf{98.5} & \textbf{86.3} & \textbf{84.1} \\
        \bottomrule
      \end{tabular}}
    \caption{Recall of gold supporting chunks on multi-hop QA samples across context lengths. All methods retrieve the top 8 chunks under an identical retrieval budget.}
    \label{tab:ablation_gating}
  \end{minipage}
\end{table*}

\subsection{Zero-shot Generalization}
\label{sec:ood}
We evaluate {\ours} zero-shot on 2WikiMultihopQA and StreamingQA. Table~\ref{tab:2WikiMultihopQA} shows strong performance on unseen multi-document reasoning tasks.  Due to space limitations, additional OOD evaluations of LongBench benchmark~\cite{bai2024longbench} on Appendix~\ref{app:ood_tasks}.

\begin{figure*}[b]
  \centering
  \captionsetup{skip=4pt}

  \begin{minipage}[t]{0.49\textwidth}
    \vspace{0pt} %
    \centering
    \includegraphics[width=\linewidth]{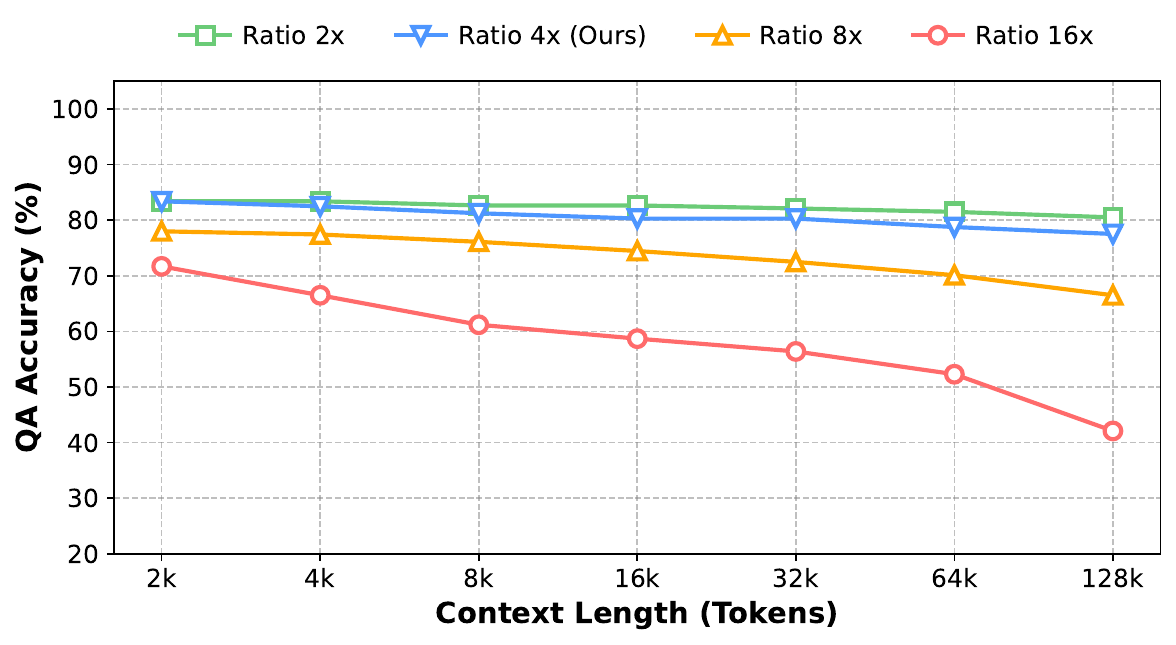}
    \caption{QA Accuracy across varying context lengths under different compression ratios. 
    The $4\times$ ratio (Ours) achieves the optimal balance, matching the stability of $2\times$ 
    while significantly outperforming aggressive compression ($16\times$).}
    \label{fig:ratio_ablation}
  \end{minipage}
  \hfill
  \begin{minipage}[t]{0.49\textwidth}
    \vspace{20pt} 
    \centering
    \small
    \resizebox{\linewidth}{!}{
      \begin{tabular}{lccc}
        \toprule
        \multirow{2}{*}{\textbf{Models / Stages}} 
        & \multicolumn{3}{c}{\textbf{Evaluation Metrics (sub-EM)}} \\
        \cmidrule(l){2-4}
        & \textbf{HotpotQA} & \textbf{2Wiki} & \textbf{Avg.} \\
        \midrule
        Qwen2.5-3B-Instruct & & & \\
        \quad $\vdash$ Stage-1: Naive Chunking & 38.28 & 35.16 & 36.72 \\
        \quad $\vdash$ Stage-2: Memory Compression & 39.84 & 36.72 & 38.28 \\
        \midrule
        \quad Stage-3: w/ SFT  & 60.16 & 58.59 & 59.38 \\
        \quad Stage-3: w/ RL   & 68.75 & 64.84 & 66.80 \\
        \quad Stage-3: w/ End-to-End RL 
          & \textbf{70.31} & \textbf{67.19} & \textbf{68.75} \\
        \bottomrule
      \end{tabular}
    }
    \vspace{10pt} 
    \captionof{table}{Ablation study of the staged optimization process. 
    The base model is Qwen2.5-3B-Instruct with a fixed context length of 16k tokens.}
    \label{tab:ablation_stages}
  \end{minipage}

\end{figure*}

\subsection{Ablation Study}
\label{sec:ablation}
To analyze the contribution of each component, we conduct a series of ablation studies using the \ours-3B model.

\subsubsection{Different Compression Ratios}
We study the effect of compression ratios ($\alpha \in \{2,4,8,16\}$) on reasoning accuracy over context lengths from 2K to 128K tokens (Figure~\ref{fig:ratio_ablation}).
Results reveal a clear trade-off between memory efficiency and information retention.
Both $2\times$ and $4\times$ compression maintain near-lossless performance, preserving $>80\%$ accuracy even at 128K, with a negligible gap ($<1\%$) between them, indicating that $4\times$ compression is sufficient to capture semantic density without redundancy.
In contrast, $16\times$ compression degrades sharply (71.5\% at 2K to 42.0\% at 128K), while $8\times$ provides a compromise but exhibits mild attrition ($<10\%$) at extreme lengths.
Accordingly, we adopt $\alpha=4$ as the default, halving the memory footprint of $\alpha=2$ with no statistically significant loss in reasoning performance.

\subsubsection{Ablation on Gate}

\paragraph{Experimental Setup.}
We evaluate different retrieval strategies under increasing context lengths by segmenting the input into non-overlapping 4096-token chunks. For the embedding baseline, we further split each 4096-token chunk into 1024-token micro-chunks, score each micro-chunk with the query, and use the maximum score as the chunk score.

\paragraph{Results and Analysis.}
As shown in Table~\ref{tab:ablation_gating}, all methods perform well at shorter contexts (56K). However, baselines show a clear performance drop as context length increases.
Static embedding-based retrieval and query-only Gate decline at 112K, with the strongest baseline dropping to 82.1\%.
In contrast, our Gate conditioned on both the query and the evolving working memory maintains a high recall of 86.3\% at 112K and 84.1\% at 224K, consistently surpassing other retrieval strategies.

This trend reflects the state-dependent nature of multi-hop reasoning: static retrievers model $P(\text{Chunk}\mid Q)$ and overemphasize early-hop evidence, whereas {\ours} conditions retrieval on the evolving memory state, modeling $P(\text{Chunk}\mid Q,\mathbf{m}_t)$, which enables adaptive evidence discovery across reasoning steps.

\subsubsection{Analysis of Staged Optimization Strategies}
Table~\ref{tab:ablation_stages} analyzes the impact of each training stage.
Memory Compression (Stage-2) achieves performance comparable to Naive Chunking (Stage-1) with reduced token usage, indicating that compression alone requires further alignment.
Stage-3 SFT yields a notable improvement (+21.10 sub-EM) by learning basic interaction patterns, but is surpassed by RL Optimization, which better supports multi-hop navigation and error correction.
The best performance (68.75 Avg. sub-EM) is obtained with End-to-End RL, where joint optimization enables gradients to reach the compressor, encouraging reasoning-aware representations and validating the need for unified perception–reasoning training.
We further provide a training convergence analysis for the joint optimization stage in Appendix~\ref{app:convergence}.

\section{Conclusion}

We introduce \ours, a cognitively inspired framework that enables efficient long-context reasoning by mimicking the human memory's division into long-term storage and dynamic working memory. Our method integrates a \texttt{Compressor}, a \texttt{Gate}, and a \texttt{Reasoner}: we jointly optimize the \texttt{Compressor} and \texttt{Reasoner} through end-to-end reinforcement learning, and train the \texttt{Gate} separately as a classifier. Experimental results demonstrate that the \ours w/o \texttt{Gate} ablation can reach up to 82\% normalized sub-EM accuracy on multi-hop benchmarks and scales context length to 1.75M tokens, while the full model provides a favorable accuracy--efficiency trade-off. Compared to MemAgent, \ours provides a 2$\times$ reduction in peak GPU memory and a 6$\times$ inference speedup. Overall, \ours offers an efficient solution for ultra-long context modeling.

\newpage

\bibliographystyle{lychee}
\bibliography{custom}

\newpage

\appendix
\appendix
\clearpage
\section{Implementation Details}
\label{app:training_details}

This appendix provides the technical specifications necessary for reproducing {\ours}. We detail the three-stage training pipeline: (1) Pre-training of the \texttt{Compressor} with \textbf{synthetic supervision} (QA pairs generated via self-annotation), (2) Joint Reinforcement Learning of the \texttt{Compressor} and \texttt{Reasoner}, and (3) \textbf{supervised} training of the \texttt{Gate} as a binary classifier.

All models are initialized from the Qwen2.5-3B-Instruct and Qwen2.5-7B-Instruct base models. We utilize 2 $\times$ NVIDIA A100 (80GB) GPUs for training.

\subsection{Stage 1: Compressor Pre-training}
\label{app:compressor_training}

The objective of the \texttt{Compressor} is to encode textual information into the latent space of memory tokens. We employ a random compression ratio $\alpha \in \{2, 4, 8, 16\}$, meaning one memory token is inserted for every $\alpha$ text tokens.

\paragraph{Data Construction.}
We sample up to 1B tokens from the RedPajama~\cite{weber2024redpajama}, and train on approximately 160M effective tokens. For each document, we create splits of sizes 2048, 4096, and 8192, denoted as $C_i$. We use self-annotation to generate synthetic Question-Answer pairs(i.e., synthetic supervision), serving as training targets for reconstruction and comprehension tasks.

\paragraph{Configuration.}
We train a LoRA adapter ($\Psi_{\text{comp}}$) for the \texttt{Compressor} with a rank of $r=64$ and a LoRA alpha of $\alpha_{\text{lora}}=128$. The larger rank is selected to ensure sufficient representation capacity for the compression task. We use the AdamW optimizer with an initial learning rate of $5\mathrm{e}{-5}$ and a cosine annealing schedule, with the maximum learning rate set to $1\mathrm{e}{-4}$. The batch size is set to 8, and training proceeds for 5,000 steps.

\subsection{Stage 2: Joint Reinforcement Learning Optimization}
\label{app:rl_training}

\begin{algorithm*}[t]
\caption{Joint Policy Optimization for {\ours} (GSPO)}
\label{alg:gspo_joint}
\begin{algorithmic}[1]
\Require Joint policy $\pi_\vartheta = \{\pi_{\text{comp}}, \pi_{\text{reason}}\}$, reference model $\pi_{\text{ref}}$ (frozen), dataset $\mathcal{D}$, group size $G$, clipping $\epsilon$, KL coefficient $\beta$
\Ensure Optimized parameters $\vartheta$
\While{not converged}
    \State Sample document--query pair $(D, Q) \sim \mathcal{D}$
    
    \For{$g = 1$ to $G$} \Comment{Group sampling for the same $(D,Q)$}
        \State Sample memory $\Theta_g \sim \pi_{\text{comp}}(\cdot \mid D)$
        \State Sample answer $A_g \sim \pi_{\text{reason}}(\cdot \mid \Theta_g, Q)$
        
        \State Define trajectory $y_g = (\Theta_g, A_g)$
        \State Compute reward $\hat{r}_g = \mathcal{R}(Q, A_g)$
        \State Compute KL penalty $d_g = D_{\mathrm{KL}}(\pi_\vartheta(y_g) \,\|\, \pi_{\text{ref}}(y_g))$
        \State $r_g \leftarrow \hat{r}_g - \beta d_g$
    \EndFor
    
    \State $\{ \hat{A}_g \}_{g=1}^G \leftarrow \textsc{GroupNorm}(\{ r_g \}_{g=1}^G)$
    
    \For{$g = 1$ to $G$}
        \State $\rho_g \leftarrow \dfrac{\pi_\vartheta(y_g)}{\pi_{\vartheta_{\text{old}}}(y_g)}$
        \State $\mathcal{J}_g \leftarrow
        \min\!\left(
        \rho_g \hat{A}_g,\,
        \text{clip}(\rho_g, 1-\epsilon, 1+\epsilon)\hat{A}_g
        \right)$
    \EndFor
    
    \State $\vartheta \leftarrow \vartheta + \eta \nabla_\vartheta
    \frac{1}{G}\sum_{g=1}^G \mathcal{J}_g$
\EndWhile
\end{algorithmic}
\end{algorithm*}

We employ the GSPO algorithm for training. The full procedure is summarized in Algorithm \ref{alg:gspo_joint}. The chunk size is set to 4096, the rollout batch size to 128, the group size $G$ to 12, and the update batch size to 16. The KL divergence coefficient $\beta$ is set to $1\mathrm{e}{-3}$. We use the AdamW optimizer. Since LoRA is the optimization target, we set the learning rate to $3\mathrm{e}{-5}$ with a linear warmup scheduler over 10 steps. In our runs, the joint optimization typically converges within 150 optimizer update steps and takes about three days of wall-clock time on 2 $\times$ A100 (80GB).

\paragraph{Reward Configuration.}
During training, we employ a strict rule-based reward validator to prevent reward hacking. We extract tokens within the \texttt{<answer></answer>} tags of the final output. If the extracted answer matches the ground truth exactly, every update step in the trajectory receives a reward of 1.0; otherwise, the reward is 0.0. We adopt this stricter validator during RL to avoid exploiting normalization artifacts that are acceptable for evaluation.

\paragraph{Dataset Construction.}
We follow the dataset construction methodology of MemAgent. Each training sample consists of 130 documents from HotpotQA, with a total token length of approximately 20K. We thoroughly cleaned the dataset by filtering out questions where Qwen2.5-7B-Instruct could achieve a 100\% Best-Of-2 score without any context (zero-shot). We selected the top 32,768 processed samples as our training set. Similarly, we synthesized 192 samples from the HotpotQA validation set. For extrapolation testing, we used the same pipeline to synthesize test sets with varying context lengths, where the number of Wikipedia entries ranges from 50 to 3,200, corresponding to context lengths from approximately 7K to 1.75 million tokens.

\subsection{Stage 3: Gate Module Training}
\label{app:gate_training}

The \texttt{Gate} is trained separately as a binary classifier to determine whether a memory block has retrieval and reasoning value given the current query and working memory.

\paragraph{Label Assignment.}
Training data is derived from the rollout process in the RL stage. For multi-hop questions, chunk updates (i.e., memory block updates) containing supporting facts are labeled as \textbf{Positive} ($y=1$). Chunk updates containing no supporting facts are labeled as \textbf{Negative} ($y=0$).

\paragraph{Objective.}
We minimize the Binary Cross-Entropy (BCE) loss. To mitigate the class imbalance problem (where irrelevant paragraphs far outnumber relevant ones), we apply a positive class weight of $pos\_weight = 3.0$.

\paragraph{Configuration.}
The Gate LoRA adapter ($\Psi_{\text{gate}}$) uses a smaller rank of $r=16$. We train for 3 epochs with a learning rate of $5\mathrm{e}{-5}$. During inference, the gating threshold $\tau$ is empirically set to 0.5.

\subsection{Evaluation and Baselines}
\label{app:eval_baselines}

\paragraph{Evaluation Metrics.}
During evaluation, we report normalized sub-EM (Exact Match). We normalize both the model answer and the ground truth (e.g., removing definite articles, ignoring case differences) and compute a sub-EM score. This means if an answer contains all elements of the standard answer, it is considered correct. When an answer consists of multiple parts, the score corresponds to the proportion of correct parts provided.

\paragraph{Long-Context Benchmarks.}
We evaluate our model on three long-context QA benchmarks, including RULER-HQA~\cite{yang2018hotpotqa, hsiehruler}, 2WikiMultihopQA~\cite{ho2020constructing}, and StreamingQA~\cite{liska2022streamingqa}. Below we describe the benchmark construction and our implementation details.
\begin{itemize}
    \item \textbf{RULER-HQA:} A synthetic long-context HotpotQA benchmark derived from the RULER framework. Similar to HotpotQA, each query has two supporting documents (gold evidence). We construct long contexts by mixing the gold evidence with irrelevant distractor documents (sourced from other samples). We build test sets with varying total context lengths ($N \in \{7\text{k}, 14\text{k}, 28\text{k}, ..., 448\text{k}, 896\text{k}, 1.75\text{M}\}$), with randomized evidence positions.
    \item \textbf{2WikiMultihopQA:} A multi-hop QA dataset built from Wikipedia. We follow the same long-context construction and evaluation pipeline as RULER-HQA: we take the query-relevant evidence documents from 2WikiMultihopQA and mix them with distractor documents to reach the target context length (14K/28K/56K in our experiments). We use the same chunking setting ($sz=4096$), memory compression, dynamic recall, and normalized sub-EM evaluation.
    \item \textbf{StreamingQA:} A streaming QA benchmark designed for evaluation under continuously growing corpora. For our long-context setting, we concatenate the documents of all questions into a single global document of approximately 800k tokens, and evaluate each question by running {\ours} over this shared 800k context. We use the same chunking setting ($sz=4096$) and normalized sub-EM evaluation.
\end{itemize}

\paragraph{Baselines.}
We compare {\ours} against three categories of strong baselines:
\begin{itemize}
    \item \textbf{RAG Agent:} We implement a standard Retrieval-Augmented Generation agent using OpenAI's \texttt{text-embedding-3-large} as the retriever. The document is segmented into semantic chunks (Wikipedia entries). For each query, the agent retrieves the top-8 most relevant chunks and feeds them into the base model for generation.
    
    \item \textbf{Search-R1:} We use a Search-R1 agent trained on Qwen2.5-7B. Similar to the RAG agent, it uses OpenAI's \texttt{text-embedding-3-large} as the retriever and retrieves the top-8 most relevant chunks. The agent then runs an iterative ReAct loop: generating a search query, retrieving context, reasoning over the results, and deciding whether to search again or answer.
    
    \item \textbf{MemAgent:} We utilize the official implementation of MemAgent \citep{yu2025memagent}. MemAgent processes long documents in fixed-size segments (set to 5k tokens by default in the official implementation). It maintains a global memory panel and employs a learnable policy to decide whether to read, write, or overwrite information at each step. We align the prompt settings and base model (Qwen2.5-7B-Instruct) with {\ours} to ensure a fair comparison of the memory mechanisms.
\end{itemize}

\subsection{Storage and Computation Trade-off}
\label{app:storage_tradeoff}

A critical challenge in long-context processing is the management of GPU memory (VRAM). Even with our efficient compression mechanism, maintaining a compressed KV-cache for extremely long sequences can impose a prohibitive storage overhead.

\paragraph{Storage Analysis.}
Consider a scenario with a context length of $N=1.75\text{M}$ tokens using the Qwen2.5-3B model (Hidden size $d=2304$, Layers $l=36$). With a compression ratio of $\alpha=4$, the system generates approximately $437.5\text{k}$ latent memory tokens. 
Since Qwen2.5-3B utilizes Grouped Query Attention (GQA) with 2 KV heads and 16 Query heads, the storage requirement for the KV-cache (in bfloat16 precision) is:
\begin{equation*}
    \mathcal{M}_{\text{KV}} \approx 2 \times l \times d_{\text{head}} \times n_{\text{kv}} \times \frac{N}{\alpha} \times 2 \text{ bytes} \approx 18.1 \text{ GB}
\end{equation*}
While this fits within the memory of high-end GPUs like the A100 (80GB), it still consumes a significant portion of VRAM, limiting the space available for activations and larger batch sizes. Furthermore, our goal is to enable efficient reasoning on consumer-grade hardware and to support scaling to even longer contexts (e.g., 10M tokens), where static storage becomes prohibitive.

\paragraph{Optimization Strategy.}
To address this, we identify two potential strategies:
\begin{itemize}
    \item \textbf{Offloading:} Temporarily offloading the compressed KV-cache to CPU RAM or NVMe SSDs and swapping them back to GPU only during the retrieval phase.
    \item \textbf{Just-in-Time (JIT) Compression:} Storing only the raw text and using the \texttt{Compressor} to regenerate the latent representations on the fly when needed.
\end{itemize}

By default, {\ours} uses offline pre-compression and stores the compressed KV-cache for inference. For contexts beyond 1M tokens, we optionally enable Just-in-Time (JIT) compression as an engineering strategy to reduce storage overhead. While re-computing embeddings incurs a computational cost, it can be more efficient than the I/O bottleneck of memory swapping. The compression process is a single parallel forward pass, whereas the reasoning process is autoregressive.
For a chunk of size 4096, compression requires only 1 forward step. In contrast, generating a reasoning chain often requires hundreds of serial steps. Therefore, the amortized computational overhead of on-the-fly compression is minimal compared to the benefits of reduced memory footprint and improved scalability.

\section{Training Convergence Analysis}
\label{app:convergence}

\begin{figure*}[t]
    \centering
    \includegraphics[width=0.9\linewidth]{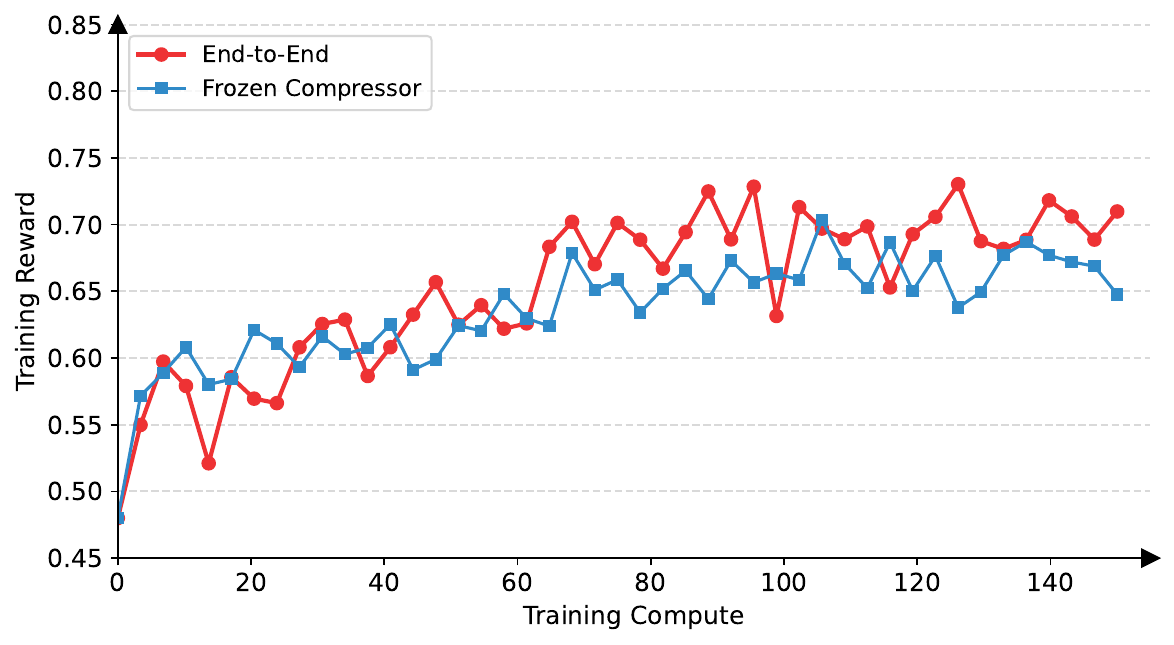}
    \caption{Training reward curves (raw data). The \textcolor{blue}{blue line} (Frozen Compressor) converges quickly but hits a performance plateau. The \textcolor{red}{red line} (End-to-End) exhibits higher variance initially due to the exploration of the compression policy but achieving higher rewards.}
    \label{fig:rl_curve}
\end{figure*}

We address the potential concern regarding the stability of jointly optimizing the Compressor and Reasoner, given the sparsity of reward signals in sequence generation tasks. Figure \ref{fig:rl_curve} presents the raw, unsmoothed training reward curves over 50 checkpoints, comparing our Joint Optimization strategy against a baseline with a Frozen Compressor.

\paragraph{Observation.} End-to-End RL curve (red) shows higher volatility in the early stages (Checkpoints 0-15) compared to the Frozen Compressor (blue). This is expected, as the gradient updates must propagate through the reasoning steps back to the compression module, causing shifts in the memory representation $\Theta$.
The Frozen Compressor rapidly converges to a local optimum ($\approx 0.67$) but fails to improve further, as the reasoner is limited by a static, suboptimal memory bank. In contrast, our method steadily climbs after the initial adaptation phase, reaching a higher reward ($\approx 0.7$).

\paragraph{Conclusion.}
The empirical results demonstrate that despite the inherent variance in RL, the joint policy successfully converges. The fluctuating but upward trend confirms that the Compressor is actively learning to retain task-critical features that maximize the Reasoner's success rate, validating the effectiveness of our end-to-end optimization framework.

\section{Failure Mode Analysis}
\label{sec:failure_analysis}

To gain deeper insights into the limitations of {\ours}, we conducted a manual error analysis on 128 randomly sampled incorrect instances from the HotpotQA and 2WikiMultihopQA validation sets. We categorized the primary causes of failure into three dominant modes: \textit{Compression-Induced Hallucination}, \textit{Unidirectional Dependency Mismatch}, and \textit{Premature Inference Anchoring}.

\subsection{Unidirectional Dependency Mismatch (35\%)}
The most significant source of error (approx. 35\%) stems from the inherent limitation of the single-pass, streaming architecture. In multi-hop reasoning, the relevance of an early piece of evidence often depends on information that appears later in the document.

\paragraph{Mechanism.}
When the model encounters a critical clue (e.g., at Step 3), it may not be semantically similar to the current query or working memory, causing the \texttt{Gate} to filter it out. Later (e.g., at Step 5), the model discovers the bridge entity that makes the previous clue relevant. However, since the static memory has already been processed and the model cannot backtrack, this information is permanently lost.

\begin{tcolorbox}[colback=red!5!white,colframe=red!50!black,title=\textbf{Case Study 1: The Late-Binding Problem}]
\small
Query: \textit{What represents the nationality of the director of the film "The Blue Kite"?}
\begin{itemize}[leftmargin=*]
    \item Step 3 (Context Chunk): "...Tian Zhuangzhuang was born in Beijing, China, and began his career..."
    \item Model Action: \textcolor{gray}{[Gate: Ignore]} $\rightarrow$ The working memory contains no link to "Tian Zhuangzhuang" yet.
    \item Step 8 (Context Chunk): "...'The Blue Kite' is a 1993 drama film directed by Tian Zhuangzhuang..."
    \item Model Action: \textcolor{blue}{[Gate: Retrieve]} $\rightarrow$ Update Working Memory: "Director is Tian Zhuangzhuang."
    \item Reasoning Failure: The model now knows the director, but the information about his nationality (China) was in Step 3, which was discarded. The model answers "Unknown" or hallucinates based on the name.
\end{itemize}
\end{tcolorbox}

\subsection{Premature Inference Anchoring (21\%)}
Approximately 21\% of errors occur when the model aggressively acts on partial evidence, forming a correct-looking but ultimately wrong conclusion early in the process. This creates a confirmation bias in the Working Memory.

\paragraph{Mechanism.}
The \texttt{Reasoner} generates an intermediate answer based on a partial match (e.g., a shared name). This incorrect entry in the Working Memory then dominates the attention mechanism, causing the model to either ignore subsequent contradictory evidence or misinterpret it to fit the existing hypothesis.

\begin{tcolorbox}[colback=orange!5!white,colframe=orange!50!black,title=\textbf{Case Study 2: Premature Anchoring}]
\small
Query: \textit{Which band's lead singer also released the solo album "Euphoria"?}
\begin{itemize}[leftmargin=*]
    \item Step 2 (Context Chunk): "...Enrique Iglesias released an album titled 'Euphoria' in 2010..."
    \item Model Action: Update Working Memory: "Candidate: Enrique Iglesias (Solo Artist)." $\rightarrow$ \textit{Wrong Path. The question asks for a band's lead singer.}
    \item Step 6 (Context Chunk): "...Def Leppard's lead singer Joe Elliott released a projected titled..." (Irrelevant text follows).
    \item Step 9 (Context Chunk): "...The band 'Morningwood' features lead singer Chantal Claret..." (Target info appears later).
    \item Reasoning Failure: The working memory is already anchored on Enrique Iglesias. The model stops actively searching for "bands" or tries to justify why Enrique fits the description, ignoring the correct entity appearing later.
\end{itemize}
\end{tcolorbox}

\subsection{Compression-Induced Hallucination (17\%)}
As analyzed in Appendix \ref{app:case}, about 17\% of errors are due to \textit{Feature Collapse} within the \texttt{Compressor}.

\paragraph{Mechanism.}
High compression ratios can cause distinct entities with similar semantic embeddings (e.g., brothers, movies in the same franchise, dates close in time) to merge in the latent space. The \texttt{Reasoner} retrieves a blurred representation, leading to attribute swapping.

\begin{tcolorbox}[colback=blue!5!white,colframe=blue!50!black,title=\textbf{Case Study 3: Attribute Swapping}]
\small
Query: \textit{Who was born earlier, William Johnson or Wilson Johnson?}
\begin{itemize}[leftmargin=*]
    \item Compressed Memory: Encodes "William... born 1856" and "Wilson... born 1860" into adjacent latent vectors.
    \item Retrieval: The Reasoner retrieves the block containing both.
    \item Reasoning Failure: Due to vector smoothing, the specific binding of dates to names is lost. The model outputs: \textit{"Wilson was born in 1856,"} effectively swapping the birth years.
\end{itemize}
\end{tcolorbox}

\subsection{Other Error Types (27\%)}
The remaining errors include:
\begin{itemize}
    \item Context Overflow: The accumulation of too many potentially relevant chunks fills the working memory context limit, flushing out early correct evidence.
    \item Instruction Misalignment: The model correctly retrieves evidence but fails to align the final answer format with user instructions (e.g., answering Yes instead of a specific entity name).
\end{itemize}

\section{Computational Complexity}
\label{app:complexity_analysis}

To rigorously evaluate the efficiency of {\ours}, we model the theoretical FLOPs required to process a document of length $N$ and generate an answer. We compare three distinct paradigms:
\paragraph{Full-Context.} Processes the entire sequence simultaneously.
\paragraph{MemAgent.} Processes the sequence in chunks, performing an autoregressive memory update for \textit{every} chunk.
\paragraph{{\ours} (Ours):} Compresses the sequence first, then employs a sparse, gated retrieval mechanism.

\subsection{FLOPs Formulation}

Let $N$ be the total document length, $sz$ be the chunk size, $L_Q$ be the query length, and $L_A$ be the output generation length. The document is segmented into $K = \lceil N/sz \rceil$ chunks. We denote the model's hidden dimension as $d$ and depth as $l$. The complexity of generating a sequence of length $L_{gen}$ given a prompt of length $L_{pmt}$ is dominated by attention, scaling as $\mathcal{O}(l \cdot d \cdot (L_{pmt} + L_{gen})^2)$.

\paragraph{Full-Context.}
The computational complexity is dominated by the global self-attention mechanism over the entire sequence.
\begin{align*}
\mathcal{C}_{\text{Full}} 
&\approx \mathcal{O}\left( l \cdot d \cdot (N + L_Q + L_A)^2 \right) \\
&\approx \mathcal{O}(N^2) \quad (\text{when } N \gg L_Q, L_A)
\end{align*}
The prefill stage processes $N + L_Q$ tokens, followed by decoding $L_A$ tokens. As $N$ grows to millions, the quadratic term makes this prohibitive.

\paragraph{MemAgent.}
MemAgent adopts a linear scanning approach but incurs a heavy constant factor due to forced memory updates. It performs generation for \textit{every} chunk.
Let the input per step be $L_{\text{in}} = L_Q + \text{Mem}_{\text{size}} + sz$, and output be $L_{\text{out}} = \text{Mem}_{\text{update}}$.
The total FLOPs sums over all $K$ steps:
\begin{align*}
\mathcal{C}_{\text{MemAgent}} 
&\approx K \times \mathcal{O}\left( l \cdot d \cdot (L_{\text{in}} + L_{\text{out}})^2 \right) \\
&= \frac{N}{sz} \times C_{\text{gen}} \approx \mathcal{O}(N)
\end{align*}
Although asymptotically linear, the constant $C_{\text{gen}}$ involves a full generation process (KV-cache read + autoregressive write) for every chunk, leading to a steep increase in computational cost.

\paragraph{{\ours} (Ours).}
Our method decouples processing into efficient compression and sparse reasoning. We utilize a compression ratio of $\alpha=4$.

\begin{itemize}
    \item Phase 1: Compression.
    The model processes chunks in parallel to encode KV-cache. Since there is no autoregressive decoding, the cost is proportional to the input tokens.
    \[ \mathcal{C}_{\text{comp}} \approx \mathcal{O}(l \cdot d \cdot N) \]
    
    \item Phase 2: Gate.
    The Gate scores all $K$ blocks. Due to compression, the effective sequence length is $N/\alpha$. The Gate requires only a single forward pass per block.
    \[ \mathcal{C}_{\text{gate}} \approx \mathcal{O}\left( l \cdot d \cdot \frac{N}{\alpha} \right) \]
    
    \item Phase 3: Reasoning.
    The \texttt{Reasoner} is activated only for the top-$T$ relevant chunks ($T \ll K$).
    \begin{align*}
    \mathcal{C}_{\text{reason}}
    &\approx T \times \mathcal{O}\left( l \cdot d \cdot (L_{\text{in}} + L_{\text{out}})^2 \right) \\
    &= T \times C_{\text{gen}}
    \end{align*}
\end{itemize}

The total cost is $\mathcal{C}_{\text{Ours}} = \mathcal{C}_{\text{comp}} + \mathcal{C}_{\text{gate}} + \mathcal{C}_{\text{reason}}$.
Comparing the dominant terms:
\begin{itemize}
    \item \textbf{MemAgent:} $\frac{N}{sz} \times C_{\text{gen}}$ (Dense Generation)
    \item \textbf{{\ours}:} $\mathcal{O}(N) + T \times C_{\text{gen}}$ (Sparse Generation)
\end{itemize}
Since $T \ll N/sz$ (e.g., retrieving only 10\% of chunks), {\ours} significantly reduces the number of expensive generation calls, resulting in a much flatter scaling curve.
\begin{figure}[t]
  \centering
  \includegraphics[width=0.95\linewidth]{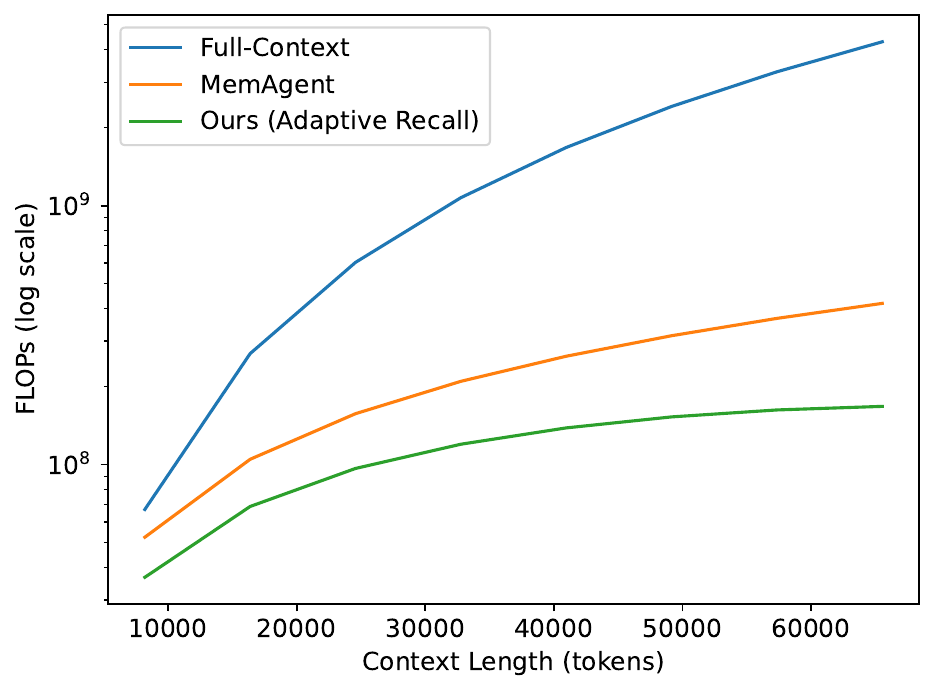} 
\caption{FLOPs Scaling Analysis (Log Scale). We compare the estimated computational cost across context lengths from 8K to 64K tokens using a logarithmic FLOPs axis. For {\ours}, the effective recall ratio is assumed to decrease linearly from 100\% to 40\% as context length increases, reflecting increasingly selective memory access under long contexts. FLOPs are analytically estimated under simplified assumptions and are intended to illustrate relative scaling trends rather than exact runtime measurements.}
  \label{fig:flops_comparison}
\end{figure}

\subsection{Quantitative Comparison}

Figure \ref{fig:flops_comparison} illustrates the FLOPs scaling behavior under increasing context lengths. The Full-Context baseline exhibits the expected $O(N^2)$ complexity due to dense self-attention, which appears as a straight line with a steep slope under the logarithmic FLOPs axis, indicating rapidly growing computational cost. In contrast, both MemAgent and {\ours} achieve linear scaling with respect to context length. Despite sharing the same asymptotic complexity, {\ours} consistently incurs lower FLOPs than MemAgent across all evaluated settings. This improvement is primarily attributed to the \textbf{Compress-then-Reason} paradigm: first, the input sequence is compressed by $4\times$, substantially reducing the number of tokens processed by the gating mechanism; second, unlike MemAgent which performs mandatory write operations for every chunk, {\ours} employs adaptive gating to activate the computationally heavy Reasoner only for a small fraction of chunks, resulting in a significantly smaller constant factor in practice.

\section{Out-of-Distribution (OOD) Generalization Analysis}
\label{app:ood_tasks}
While {\ours} is primarily optimized for multi-hop QA, we also test whether its compressed memory bank $\Theta$ transfers to other long-context formats. We conduct OOD experiments on long-document summarization tasks from LongBench~\citep{bai2024longbench}, which differ from the QA-based training setup. We consider GovReport (summarizing lengthy government reports) and MultiNews (summarizing multiple news documents). We report ROUGE-1 (R-1), ROUGE-2 (R-2), and ROUGE-L (R-L) scores, and compare our method against the base model Qwen2.5-7B-Instruct as well as MemAgent-7B.

\begin{table}[t]
\centering
\small
\renewcommand{\arraystretch}{1.2}
\setlength{\tabcolsep}{5pt}
\begin{adjustbox}{width=1.0\linewidth}
\begin{tabular}{l|cccc|cccc}
\toprule
\multirow{2}{*}{\textbf{Model}} & \multicolumn{4}{c|}{\textbf{GovReport}} & \multicolumn{4}{c}{\textbf{MultiNews}} \\
\cmidrule(lr){2-5} \cmidrule(lr){6-9}
 & \textbf{R-1} & \textbf{R-2} & \textbf{R-L} & \textbf{Avg.} & \textbf{R-1} & \textbf{R-2} & \textbf{R-L} & \textbf{Avg.} \\
\midrule
Qwen2.5-7B-Instruct & \textbf{30.91} & 11.68 & 15.20 & 19.26 & 46.64 & 12.01 & 28.33 & 28.99 \\
MemAgent-7B & 30.28 & 12.37 & \textbf{15.37} & 19.34 & \textbf{48.49} & 14.41 & \textbf{30.91} & \textbf{31.27} \\
\textbf{\ours-7B} (ours) & 30.12 & \textbf{13.08} & 15.07 & \textbf{19.42} & 47.43 & \textbf{15.28} & 30.33 & 31.01 \\
\bottomrule
\end{tabular}
\end{adjustbox}
\caption{Zero-shot performance on LongBench summarization tasks. Despite being trained for QA, {\ours} achieves competitive performance compared to MemAgent, demonstrating strong generalization capabilities.}
\label{tab:ood_summarization}
\end{table}

\paragraph{Results and Discussion.}
As shown in Table \ref{tab:ood_summarization}, {\ours} exhibits strong zero-shot generalization. These results suggest that the compressed memory bank $\Theta$ constructed by {\ours} captures transferable semantic features that support both targeted extraction (QA) and global aggregation (summarization).

\section{Ablation on Working Memory Capacity}
\label{app:wm_ablation}

In the dynamic recall and reasoning phase, the capacity of the Working Memory ($\mathbf{m}$) is a critical hyperparameter. Given that we segment documents into chunks of size $L_{\text{chunk}} = 4096$, the working memory capacity determines the maximum buffer size available for the active context before synthesis. We investigate the impact of varying the working memory capacity limit ($L_{\text{WM}} \in \{1024, 2048, 3072, 4096\}$) on the RULER-HQA benchmark across increasing context lengths ranging from 7K to 56K tokens.

\begin{table}[t]
\centering
\renewcommand{\arraystretch}{1.2}
\setlength{\tabcolsep}{8pt}
\begin{adjustbox}{scale=1}
\begin{tabular}{c|cccc|c}
\toprule
\textbf{WM Capacity} & \multicolumn{4}{c|}{\textbf{RULER-HQA Context Length}} & \textbf{Avg.} \\
($L_{\text{WM}}$) & \textbf{7K} & \textbf{14K} & \textbf{28K} & \textbf{56K} & \textbf{Score} \\
\midrule
1024 & 82.03 & 79.69 & 78.91 & 77.34 & 79.49 \\
2048 & 82.03 & 78.91 & 78.91 & 77.34 & 79.30 \\
3072 & 81.25 & 78.91 & 77.34 & 75.78 & 78.32 \\
4096 & 80.47 & 77.34 & 76.56 & 75.00 & 77.34 \\
\bottomrule
\end{tabular}
\end{adjustbox}
\caption{Ablation results on Working Memory Capacity. The standard capacity of 1024 achieves the best performance, indicating that larger buffers do not necessarily improve reasoning.}
\label{tab:wm_ablation}
\end{table}

\paragraph{Results and Analysis.}
To ensure fairness and isolate the effect of capacity, we excluded the Gate mechanism in this ablation, as it was specifically optimized for 4096-token chunks. As shown in Table \ref{tab:wm_ablation}, we observe that performance does not improve with increased working memory capacity; in fact, $L_{\text{WM}}=1024$ yields the highest average accuracy of 79.49\%. The results suggest that expanding the working memory budget beyond the necessary chunk size tends to introduce excessive noise tokens. This accumulated noise distracts the model's attention mechanism during the final answer synthesis, leading to a degradation in reasoning precision rather than an improvement.

\section{Dynamic Evolution of Working Memory}
\label{app:memory_length}

\begin{figure}[t]
    \centering
    \includegraphics[width=\linewidth]{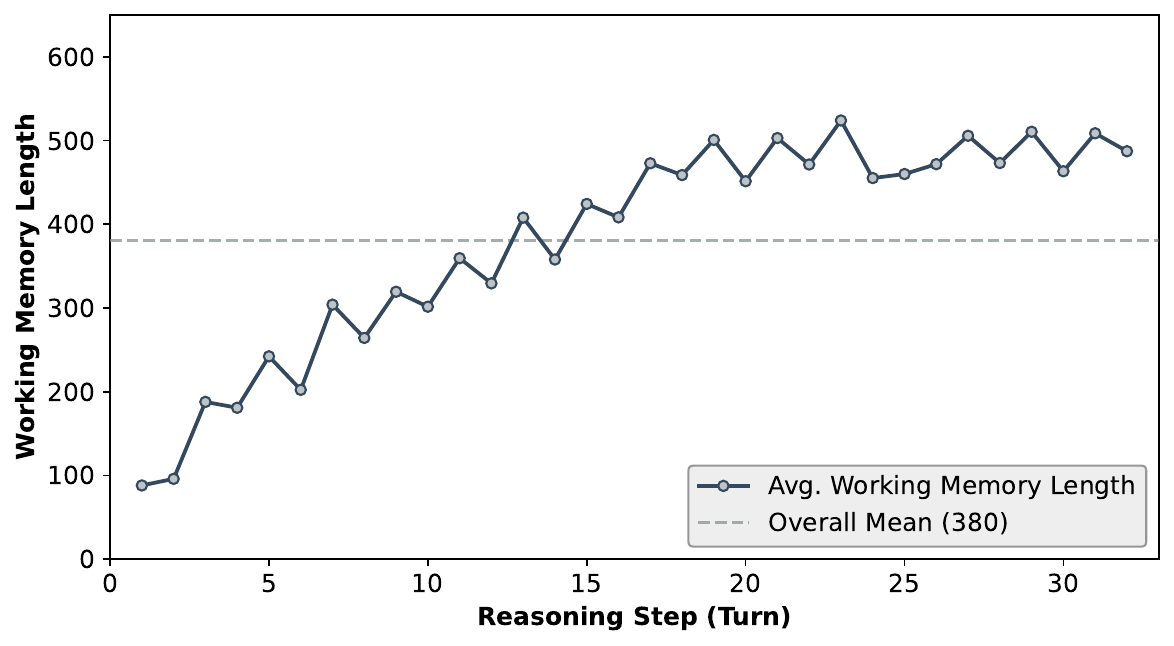}
    \caption{Evolution of Working Memory length. The length grows initially as evidence is collected but stabilizes at a peak of $\approx$ 500 tokens around Turn 23. This demonstrates that {\ours} effectively manages its context budget without unbounded growth.}
    \label{fig:memory_length}
\end{figure}

To verify the efficiency of our reasoning mechanism, we tracked the token usage of the Working Memory state $\mathbf{m}$ across 32 reasoning steps. Figure \ref{fig:memory_length} presents the average length evolution on 128 samples from the 128K context validation set.

\paragraph{Analysis.}
As shown in Figure \ref{fig:memory_length}, the working memory length exhibits a clear saturation behavior rather than monotonically increasing with reasoning steps. While the memory expands during the early stages to accumulate relevant evidence, it gradually stabilizes at approximately 503 tokens—well below the predefined capacity limit of 1024. This indicates that the model does not passively append retrieved information, but instead learns to actively regulate its memory state by retaining only task-critical content. Such learned memory management effectively prevents unbounded context growth and ensures stable, efficient reasoning over long interaction horizons.

\section{Case Analysis}
\subsection{Hallucination via Feature Collapse}
\label{app:case}
To understand the failure mode of high compression, we conduct a qualitative analysis using a constructed synthetic narrative, \textit{Chronicle of the Four Johnson Brothers}, which is dense with confounding entities (similar names and dates). We compare the generated responses of the $4\times$ (Ours) and $16\times$ models.

\begin{table}[t]
    \centering
    \small
    \begin{tabular}{p{0.95\linewidth}}
    \toprule
    Query: Distinguish between the wives of William Henry Johnson and Wilson Harold Johnson. \\
    \midrule
    Context Facts: \\
    1. William's wife: Elizabeth Ann Smith, born \textbf{1860}. \\
    2. Wilson's wife: Elizabeth Marie Smith, born \textbf{1865}. \\
    \midrule
    Model Response ($\alpha=16$, High Compression): \\
    "William's wife was Elizabeth Ann... Wilson's wife was Elizabeth Marie... \textcolor{red}{Both women were born in 1860}, but their names are distinct." \\
    \textit{(Error: Hallucinated Wilson's wife's birth year by merging it with William's wife's.)} \\
    \midrule
    Model Response ($\alpha=4$): \\
    "Elizabeth Ann was born in \textbf{1860}, while Elizabeth Marie was born in \textbf{1865}. They are cousins." \\
    \bottomrule
    \end{tabular}
    \caption{Case study on fine-grained information retrieval. The $16\times$ model suffers from attribute merging (hallucination), while $4\times$ retains precision.}
    \label{tab:case_study}
\end{table}

\paragraph{Discussion.}
As shown in Table \ref{tab:case_study}, while the $16\times$ model correctly retrieves high-level entities (names), it fails at \emph{attribute binding}. It incorrectly assigns the birth year ``1860'' to both wives. This suggests that aggressive compression causes \emph{feature collapse} in the latent space, where distinct numerical tokens (1860 vs. 1865) closer in proximity are averaged into a single representation. In multi-step reasoning, these small hallucinations accumulate, leading to the rapid performance decay observed in the $16\times$ curve (Figure \ref{fig:ratio_ablation}). The $2\times$, $4\times$, and $8\times$ models correctly identified the dates in this test, confirming that moderate compression preserves the fidelity required to distinguish fine-grained details.

\subsection{Associative Reasoning and Self-Correction}
\begin{tcolorbox}[casebox, title=Question: The city that contains the Yunnan Provincial Museum is also known by what nickname?]
    
    \textbf{\textcolor{seedblue}{Chunk k}}
    
    ...
    The Yunnan Provincial Museum is located in Kunming, the capital city of Yunnan Province in southwest China.
    The museum features exhibitions on Yunnan's history, ethnic groups, and archaeology.
    ...
    \tcbline 
    
    \textbf{\textcolor{seedblue}{Working memory k:}}

    - The Yunnan Provincial Museum is located in Kunming City, Yunnan.
    
    - Yangzonghai Lake is located near Kunming, Yunnan, in southern China.
    
    \tcbline 
    \textbf{\textcolor{seedblue}{Chunk k+1}}
    
    ...
    Dian Lake is a large freshwater lake on the Yunnan--Guizhou Plateau near Kunming.
    Dian Lake is sometimes described as the ``Sparkling Pearl Embedded in a Highland''.
    ...
    \tcbline 
    
    \textbf{\textcolor{seedblue}{Working memory k+1:}}

    - The Yunnan Provincial Museum is located in Kunming City, Yunnan.
    
    - Dian Lake is a large lake located on the Yunnan-Guizhou Plateau close to Kunming, Yunnan, its nickname is ``Sparkling Pearl Embedded in a Highland''.
    
    - Yangzonghai Lake is located near Kunming, Yunnan, in southern China.
    
    \textbf{The city that contains the Yunnan Provincial Museum might be ``Sparkling Pearl Embedded in a Highland''.}
    
    \tcbline 
    \textbf{\textcolor{seedblue}{Chunk k+2}}
    
    ...
    Kunming has a mild climate throughout the year and is widely known as the ``Spring City''.
    The city serves as a major cultural and transportation hub in Yunnan.
    ...
    \tcbline 
    
    \textbf{\textcolor{seedblue}{Working memory k+2:}}

    - The Yunnan Provincial Museum is located in Kunming City, Yunnan.
    
    - Dian Lake is a large lake located on the Yunnan-Guizhou Plateau close to Kunming, Yunnan, its nickname is ``Sparkling Pearl Embedded in a Highland''.

    - Kunming is also known as the ``Spring City'' due to its weather.
    
    - Yangzonghai Lake is located near Kunming, Yunnan, in southern China.
    
    \textbf{The new content provides Kunming's exact nickname. Therefore, the city that contains the Yunnan Provincial Museum is known as the ``Spring City''.} 
\label{fig:case_study_box}
\end{tcolorbox}
As illustrated in the case study above, the model demonstrates robust \emph{associative reasoning} and \emph{self-correction} capabilities.
Initially, the agent identifies ``Kunming'' as the pivotal entity by associating the Yunnan Provincial Museum with its location (Chunk $k$). 
It then forms a tentative hypothesis after processing memory block $k+1$, incorrectly inferring that the city's nickname might be ``Sparkling Pearl'' based on the proximal descriptions of Dian Lake. 
However, the reasoning remains flexible; upon retrieving memory block $k+2$ which explicitly describes Kunming as the "Spring City," the agent successfully detects the conflict. 
It differentiates the distraction (the lake's nickname) from the city's actual alias and rectifies its working memory, effectively overriding the previous tentative inference with the verified fact.

\section{Comparison with  RAG}
\label{app:rag_comparison}

RAG and {\ours} address different bottlenecks in long-context reasoning. RAG enables fast retrieval with high recall via approximate nearest neighbor search, making it well-suited for large external corpora. Our goal is not to replace RAG, but to provide an alternative long-context processing paradigm based on compressed memory and state-dependent retrieval: RAG typically ranks chunks by query-only similarity and may miss late-hop evidence that becomes relevant only after intermediate entities are discovered, while {\ours} conditions the \texttt{Gate} on both the query and the evolving working memory $\mathbf{m}$ (Table~\ref{tab:main_results}). The two approaches are complementary: documents retrieved by RAG can be treated as additional context streams, compressed into $\Theta$, and then reasoned over by the same dynamic recall and reasoning workflow.

\end{document}